\documentclass[letterpaper]{article} 
\usepackage{aaai25}  
\usepackage{times}  
\usepackage{helvet}  
\usepackage{courier}  
\usepackage[hyphens]{url}  
\usepackage{graphicx} 
\usepackage{natbib}  
\usepackage{caption} 
\usepackage{algorithm}
\usepackage{algorithmic}

\usepackage{amsmath}
\usepackage{amssymb}
\usepackage{amsfonts}
\usepackage{multirow}
\usepackage{subfig}

\usepackage{booktabs}

\usepackage{newfloat}
\usepackage{listings}
\DeclareCaptionStyle{ruled}{labelfont=normalfont,labelsep=colon,strut=off} 
\floatstyle{ruled}
\newfloat{listing}{tb}{lst}{}
\floatname{listing}{Listing}
\usepackage{xspace}

\newcommand{\Name}{\texttt{FedBSS}\xspace}

\title{Federated Learning with Sample-level Client Drift Mitigation}
\author{
    Haoran Xu\textsuperscript{\rm 1}\equalcontrib,
    Jiaze Li\textsuperscript{\rm 1}\equalcontrib,
    Wanyi Wu\textsuperscript{\rm 2},
    Hao Ren\textsuperscript{\rm 3}\thanks{The corresponding author.}
}
\affiliations{
    \textsuperscript{\rm 1}Zhejiang University, Hangzhou, China\\
    \textsuperscript{\rm 2}Shandong University, China\\
    \textsuperscript{\rm 3}School of Cyber Science and Engineering, Sichuan University, Chengdu 610065, China\\
    xhaoran@zju.edu.cn, Jiaze\_Li@zju.edu.cn,\\
    wanyi\_wu@mail.sdu.edu.cn,
    hao.ren@scu.edu.cn

}

\usepackage{bibentry}

\begin{document}

\maketitle

\begin{abstract}
Federated Learning (FL) suffers from severe performance degradation due to the data heterogeneity among clients. Existing works reveal that the fundamental reason is that data heterogeneity can cause \textit{client drift} where the local model update deviates from the global one, and thus they usually tackle this problem from the perspective of calibrating the obtained local update. Despite effectiveness, existing methods substantially lack a deep understanding of how heterogeneous data samples contribute to the formation of client drift. 
In this paper, we bridge this gap by identifying that the drift can be viewed as a cumulative manifestation of \textit{biases} present in all local samples and the bias between samples is different. Besides, the bias dynamically changes as the FL training progresses. 
Motivated by this, we propose \Name that first mitigates the heterogeneity issue in a sample-level manner, orthogonal to existing methods. Specifically, the core idea of our method is to adopt a \underline{b}ias-aware \underline{s}ample \underline{s}election scheme that dynamically selects the samples from small biases to large epoch by epoch to train progressively the local model in each round. In order to ensure the stability of training, we set the diversified knowledge acquisition stage as the warm-up stage to avoid the local optimality caused by knowledge deviation in the early stage of the model. Evaluation results show that \Name outperforms state-of-the-art baselines. In addition, we also achieved effective results on feature distribution skew and noise label dataset setting, which proves that \Name can not only reduce heterogeneity, but also has scalability and robustness.
\end{abstract}

%

\section{Introduction} 
Federated Learning (FL), which enables collaborative training of models by sharing parameters among clients without exchanging raw data, has garnered significant attention for its ability to leverage vast amounts of data on clients while preserving privacy. 
The basic steps of FL are to iteratively execute local model training on multiple clients individually and subsequently aggregate all updated models at the server. Despite its implementation simplicity, the challenge arises from the inherently heterogeneous distribution of FL's training data across clients—characterized by non-identical and independent (Non-IID) data—substantially deteriorating the performance of the obtained global model.

Many approaches have been devoted to addressing the data heterogeneity problem of FL. 
One of the representative categories is to consider that the essence of performance degradation is due to the \textit{client drift} caused by the data heterogeneity where the local updates of each client greatly deviate from the aggregated global one~\citep{pmlr-v119-karimireddy20a,DBLP:conf/mlsys/LiSZSTS20,pmlr-v162-zhang22p,10.1145/3447548.3467254,gao2022federated,louizos2024mutual}.
To mitigate this drift, many methods are proposed to calibrate the local update from the perspective of optimization ~\citep{gao2022federated,pmlr-v119-karimireddy20a}. For example, some methods leverages the difference between the local update and the global update in old rounds to compensate the local update in current round~\citep{pmlr-v119-karimireddy20a,gao2022federated} and other works add regularization on the local loss function to facilitate the local update to approach the global one~\citep{pmlr-v162-zhang22p,10.1145/3447548.3467254,louizos2024mutual}. Besides a few works explore different aggregation strategies on the server~\citep{chan2024internal}.
\par
Although these approaches have made great achievements, they lack a deep understanding of \textbf{\textit{how the heterogeneous data samples contribute to the formation of client drift}}.

In this paper, we seek to tackle the Non-IID challenge of FL from the sample level by delving into the impact of each local sample on the drift of local update. 
Specifically, we identify that \textbf{there exists substantial bias in each sample and the client drift can be viewed as a cumulative manifestation of the biases of all samples}. Besides, the bias between samples is different from each other where the sample with large loss is considered to have a relatively large impact on the client drift. Also, we find that the bias of each sample dynamically changes during the FL training process. 
%
\par
Motivated by this finding, we propose a simple yet effective method, \Name, which includes two stages: Diversified knowledge acquisition stage and Progressive knowledge learning stage.
The diversified knowledge acquisition phase acts as a model's cold start, warming it up by introducing varied knowledge through standard aggregation techniques. This step prevents early-stage deviations caused by limited knowledge scope, essentially initializing the global model with a diverse set of preliminary information guided by the principle of diversity.
The second stage is to dynamically select samples based on their bias to train the local model. Specifically, we measure the bias of each sample by the loss of the global model. Then, during each round, each client sorts the bias of all samples and selects the samples with small loss to train the local model. 
Considering that filtering out samples with large biases may reduce the Knowledge acquisition efficiency, we further propose progressively adding samples with large biases to the training process.
Finally, in order to give the partition boundary of samples with different biases adaptively, we introduce the concept of uncertainty.
%
%
\par
Experiment results on different datasets and models show that \Name can significantly improve the performance of the global model under label distribution skew, feature distribution skew and noise data settings compared to baselines. Our contributions can be summarized as:

\begin{itemize}
\item To the best of our knowledge, this paper is the \textit{first} to consider to solve client drift issues in the sample level. Our findings reveal that the biases of client samples are different from each other which cumulatively formalizes the client drift. 
%

\item We propose a novel and fine-grained two-stage approach, \Name, which has the diversified knowledge acquisition stage and the progressive  knowledge learning stage. The former is based on the principle of diversity and infuses diversified knowledge into the global model.
On the basis of the former, the latter sorts the loss of client sample based on the current global model, adaptively divides the client sample by uncertainty, and guides the global model from unbiased sample to biased samples learning through progressive knowledge learning strategies.

\item To validate the efficiency of the proposed method, we compare \Name with state-of-the-art methods on Non-IID data settings. By conducting extensive experiments over various deep-learning models and datasets, we show that \Name can not only reduce data heterogeneity but also has scalability and robustness. 
\end{itemize}
\par
\begin{figure*}[t]
\centering
\includegraphics[width=1 \linewidth]{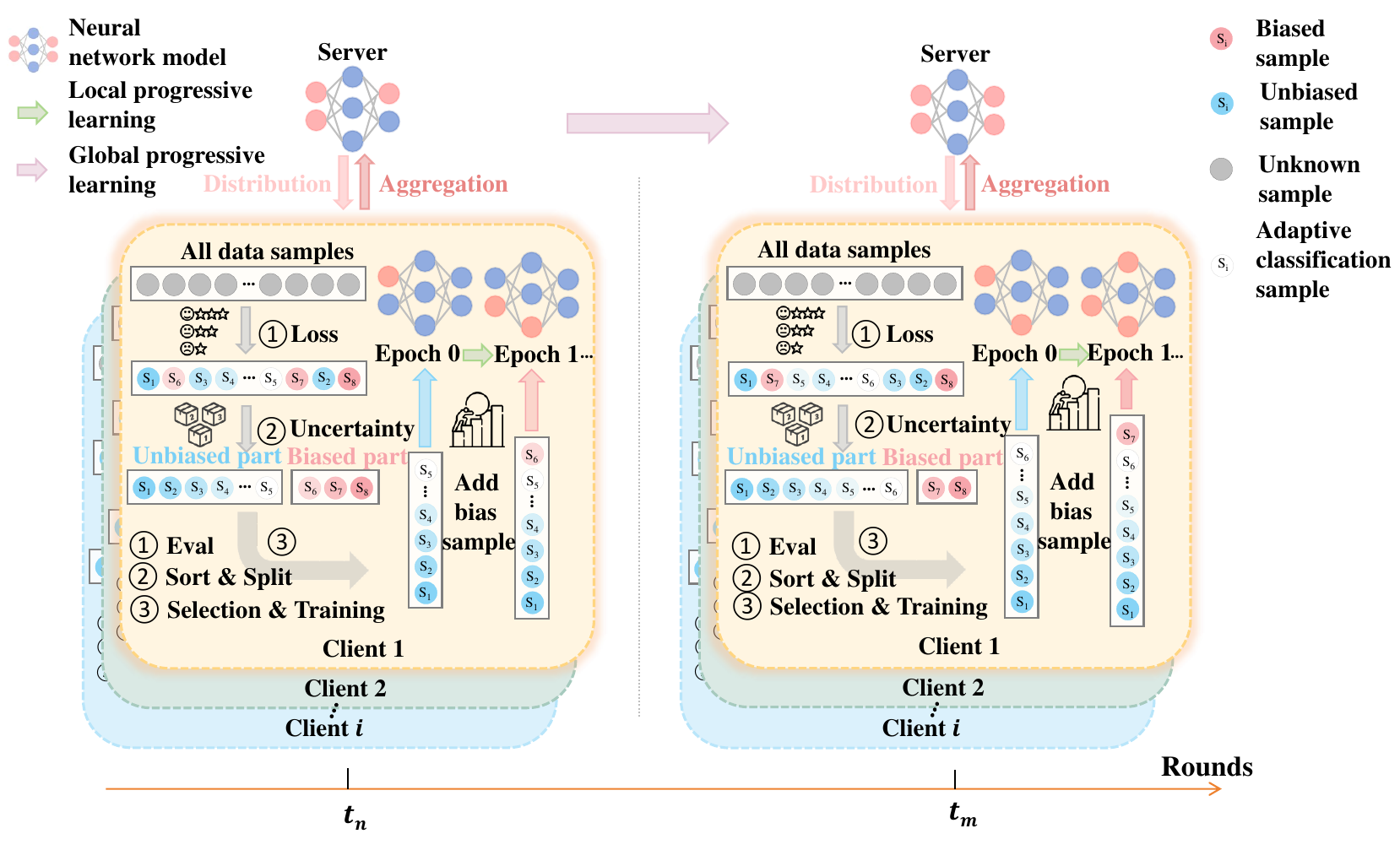}
\caption{Illustration of \Name. Our approach includes local and global progressive learning, shown by green and pink arrows. In the client's local learning, there are three steps: 1) Samples are rated by loss; higher $i$ in $S_i$ signifies greater loss (redder), lower means lesser (bluer). 2) Loss-based sorting identifies an adaptive threshold for classifying samples into 'unbiased' (low loss) or 'bias' (higher loss) sets. 3) Training begins with 'unbiased' samples, then integrates 'bias' ones over time, enabling comprehensive learning.
Meanwhile, the global model learns biases across rounds, shrinking the 'bias' set.
}
\label{fig:pipeline} 
\end{figure*}

\section{Related Works}\label{sec:related}
Many approaches~\citep{li2021model,pmlr-v139-zhu21b,NEURIPS2022_fadec8f2,acar2021federatedlearningbaseddynamic} have been dedicated to addressing the issue of data heterogeneity in Federated Learning (FL). These methods can be classified into two primary categories: those that calibrate local updates from an optimization standpoint and those that devise aggregation strategies. The specifics of these related works are elaborated upon as follows.

\textbf{Calibrating Local Update} 
Previous research has tackled the challenge of Non-IID data in federated learning by calibrating the local model updates. Works such as SCAFFOLD~\citep{pmlr-v119-karimireddy20a} address client drift by employing control variates for gradients, albeit without directly considering inconsistencies between local and global objectives. FedDC~\citep{gao2022federated}, in a different approach, utilizes auxiliary drift variables to monitor and mitigate discrepancies between local and global model parameters.
Regarding statistical heterogeneity, FedProx~\citep{DBLP:conf/mlsys/LiSZSTS20} integrates an additional term into the local model's objective function, suggesting that over-updating the local model could hinder global convergence. This proximal term serves as a penalty, discouraging significant deviations of the local model from the global one. Another strategy, presented by FedLC~\citep{pmlr-v162-zhang22p}, introduces a fine-tuned calibrated cross-entropy loss for local updates. It does so by incorporating a pairwise label margin, enhancing the sensitivity of the learning process to inter-class differences.
Similarly, FedRS~\citep{10.1145/3447548.3467254} addresses the challenge of missing classes by proposing a "Restricted Softmax" mechanism, which constrains the updating of weights associated with missing classes during local training. FedDyn~\citep{acar2021federated}, on the other hand, introduces a dynamic regularizer tailored to each client to align individual client models with the global model, thereby reducing communication overhead. More recently, integrating contrastive learning and mutual information into the loss function has emerged as another optimization strategy, as illustrated by Louizos et al.~\citep{louizos2024mutual}. Strategies such as incorporating additional regularizers into loss functions or leveraging gradient control mechanisms are all aimed at mitigating the drift of local models.

\textbf{Aggregation Strategy} 
The parameters of client models are inherently prone to drift, making server aggregation strategies another pivotal approach to mitigating the effects of Non-IID data. One such strategy, InCo ~\citep{chan2024internal}Aggregation, harnesses internal cross-layer gradients—combining gradients from both shallow and deep layers within a server model—to enhance similarity in the deeper layers without necessitating extra client-server communication.Recently, FedCDA~\citep{wang2024fedcda}, on the other hand, employs a selective aggregation of cross-round local models, effectively minimizing disparities between the global and local models.
%
\section{Problem Formulation}\label{sec:preliminaries}
In this setting, we have $N$ clients with their private datasets $\mathcal{D}_{n}=\left\{\left(\boldsymbol{x}_{i}, y_{i}\right)\right\}_{i=1}^{S_{n}}$ where ${x}_{i}$ is the data sample, ${y}_{i}$ is its label and $S_{n}$ is the number of training sample on $n$-th client. The objective of federated learning framework is to learn the global model parameter $\vartheta$ which minimizes the loss function $F(\vartheta)$ on training data of all clients without access to original data:
\begin{equation}\label{eq:obj}
\mathop{min}\limits_{\vartheta \in \mathbb{R}^d} F(\vartheta) = \frac{1}{N}\sum_{n=1}^{N}F_n(\vartheta)
\end{equation}
\begin{equation}\label{eq:localObj}
    F_n(\vartheta) = \mathbb{E}_{\boldsymbol{x} \sim \mathcal{D}_n}f_n(\vartheta;\boldsymbol{x})
\end{equation}
$f_n(\vartheta;\boldsymbol{x})$ denotes the loss value with respect to model $\vartheta$ and random data sample $\boldsymbol{x}$.

\section{Motivation}\label{sec:motivation}
In this section, we revisit the principles of client drift and ask three questions:

\textbf{\text {Q}1: What is the impact of different local samples on client drift?}
\par
\begin{figure}
    \centering
     \includegraphics[width=1 \linewidth]{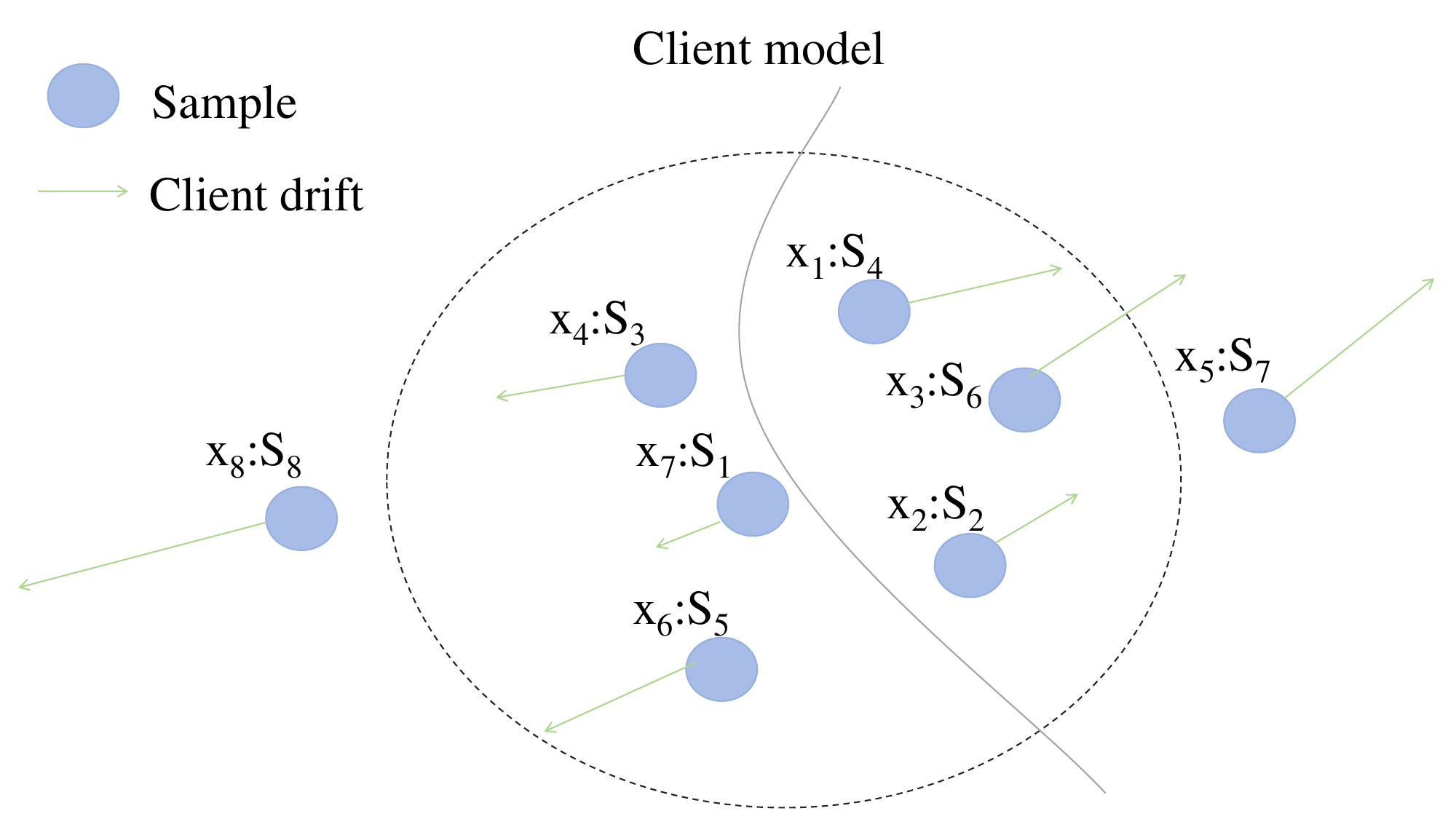}
    \caption{Different samples on client have a different degree of drift to the model.}
\label{fig:motivation}
\end{figure}
\begin{figure*}[!h]
    \centering
    \subfloat[Visualization different drift]{
    \includegraphics[width=0.24\linewidth]{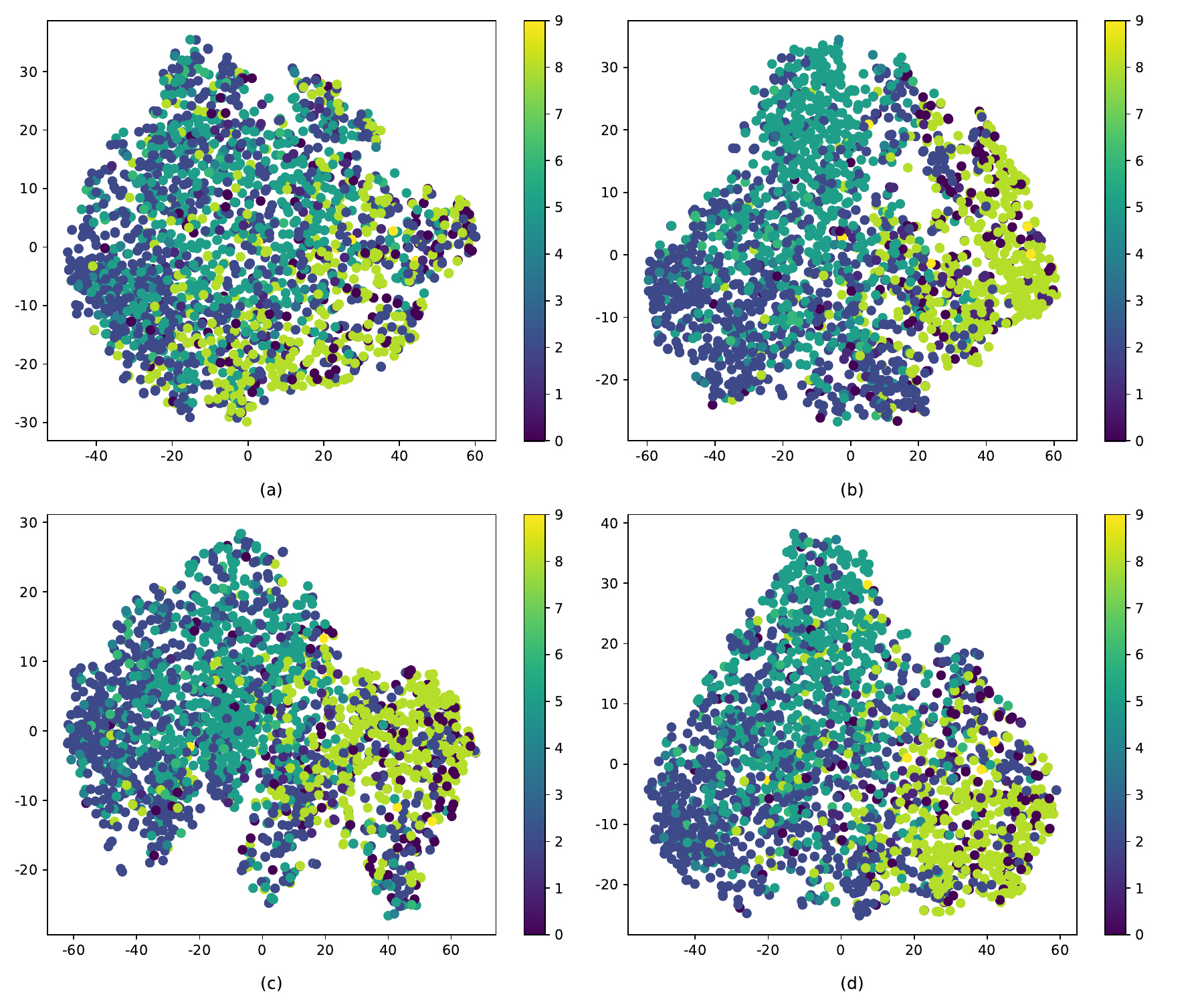}
    \label{fig:different}
    }
    \subfloat[Loss and Uncertainty]{
    \includegraphics[width=0.24\linewidth]{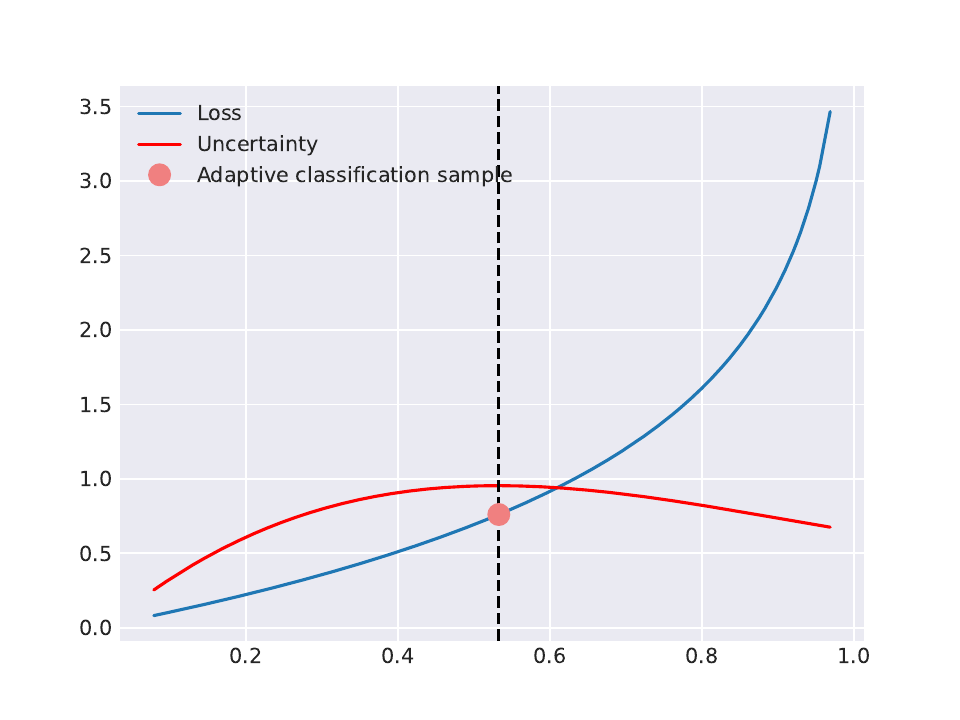}
    \label{fig:uncertainty}
    }
    \subfloat[Uncertainty samples ]{
    \includegraphics[width=0.24\linewidth]{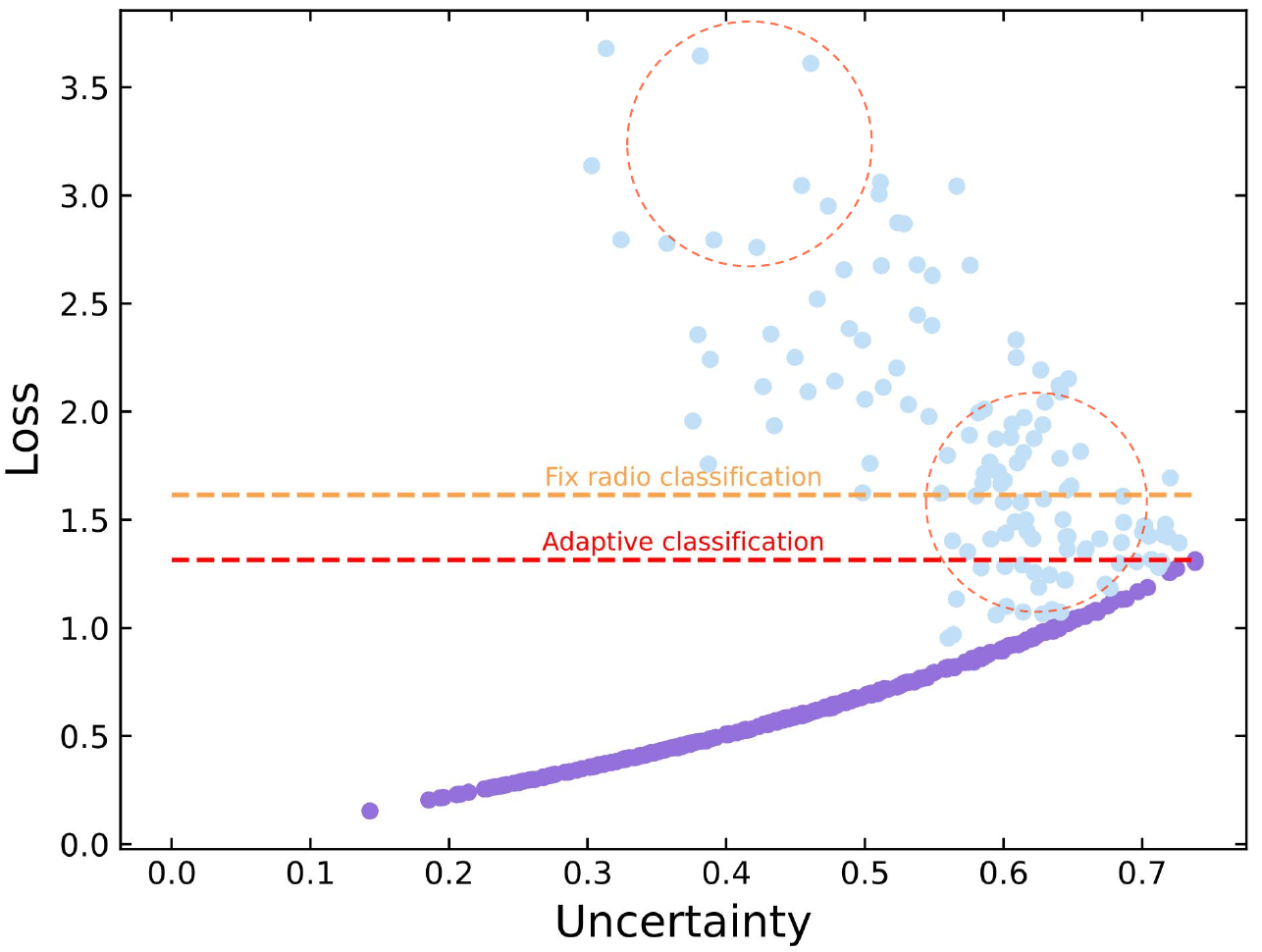}
    \label{fig:uncertaintysample} 
    }
    \subfloat[Variance of uncertainty]{
    \includegraphics[width=0.24\linewidth]{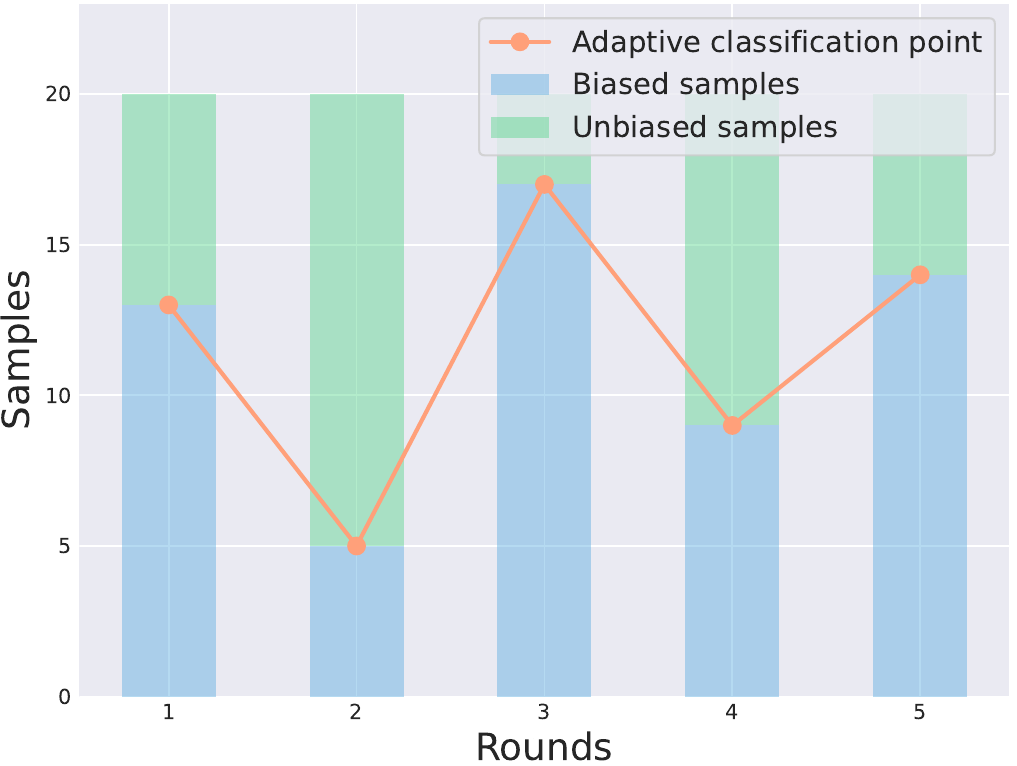}
    \label{fig:fluctuate} 
    }
    \caption{(a) Different impacts of various local samples on model drift. (b) The relationship between loss and uncertainty changes. (c) The relationship between uncertainty and loss as local samples vary. (d) The abrupt changes in the adaptive classification points in each round when a diversified knowledge acquisition stage is absent.}
  \label{Fig:data_heterogeneity} 
\end{figure*}

This question stems from that previous studies did not study the impact of local samples on client drift. We investigate this by profiling the training process of clients over local client samples and identify that different client samples have different drift degrees to the training of the local model. Figure \ref{fig:motivation} shows the above motivation. In fact, this naturally holds because the loss of the global model is different for each client sample where loss is a measure of the gap between the knowledge of local data and the knowledge of the global data. Therefore, we take loss value as the basis to measure the degree of the sample's influence on the client drift. To verify the above viewpoint, we track the training progress of different samples on a single client with ResNet-18~\cite{he_deep_2016} model as an example. Based on loss, we sort client samples and extract the top 50\% sample, the bottom 50\% sample and all the samples  respectively for training. Figure \ref{fig:different} is divided into four parts, which are the origin sample, 100\% sample training, top 50\% sample training and bottom 50\% sample training according to the order of top left, top right, bottom left and bottom right. We can draw the following conclusion: the degree of the client model drift is also small when the loss of samples is small and vice versa. \textbf{\textit{Therefore, we can partition the client sample into two sets based on the loss value, namely, the set of biased samples and unbiased samples which indicates samples contribute large and small to the degree of model drift separately.}}

\textbf{\text {Q}2: How to identify biased and unbiased samples from local samples?}
\par
A natural observation is that the loss of each sample in different rounds varies, causing a fixed loss threshold or ratio-based biased sample identification scheme to not apply to the scenario, which can also be observed from Figure~\ref{fig:uncertaintysample}. To solve this, inspired by previous work~\citep{fuchsgruber2024uncertaintyactivelearninggraphs,zhu-etal-2008-active} \textbf{\textit{we identify that the uncertainty, i.e., the confidence of the model on the probability of sample classification
, can adaptively give the partition boundary of unbiased samples and biased samples.}}


\textbf{\text {Q}3: How to mitigate the client drift with identified sample bias?}
\par
The core idea is that unbiased samples can reduce the drift degree of the model and increase the stability of training compared with biased samples. The intuitive idea is to filter out the biased sample part during client model training and only adopt the unbiased samples. However, the limit is that biased samples, as a part of all training samples, often contain information necessary for model training. Directly filtering out biased samples for training will affect the model's knowledge acquisition of this part of the data, and ultimately affect the model's performance. 
\textbf{\textit{To this end, the method should satisfy two requirements: 1. minimized bias to reduce client drift; 2. minimized knowledge sacrification.}}
Built upon the above analysis, we introduce a progressive learning concept, which gradually adds biased samples to the local training process. The intuition behind this is that when we gradually add biased samples, we can gradually guide the model from local to global stability to learn the knowledge in biased samples.

\section{Methodology}\label{sec:method}

\begin{algorithm}[tb]
\caption{\Name Algorithm Stage 2}
\label{alg:1}
\renewcommand{\algorithmicrequire}{\textbf{Input:}}
    \renewcommand{\algorithmicensure}{\textbf{Output:}}
\begin{algorithmic}[1]
    \REQUIRE Global model $\vartheta$, and learning rate $\eta$, stage 1 communication rounds $T_1$, stage 2 communication rounds $T_2$, total communication rounds $T_1 + T_2$.
	\ENSURE Trained global model $\vartheta$.
        

    \STATE Receive ${\vartheta}_{T_1}$ from the stage 1;
    \FOR {round $t \in \{T_1+1,T_1+2,...,T_1+T_2\}$}
        \STATE Randomly select a subset of clients $\mathcal{N}_t$;
        \STATE Distribute $\vartheta_{t-1}$ to each selected client as $\vartheta_{t-1}^n$;
    	\FOR {each selected client $n$ \textbf{in parallel}}
            \FOR {local epoch $e=1$ to $e_{total}$ \textbf{on client}}
    		\STATE $\mathbf{X}_{t,e}^{n} \leftarrow$ $\mathbf{X}_{t,e}^{n}$ update use Eq (\ref{eq:progressive});
            \STATE $\vartheta_{t,e+1}^n = \vartheta_{t,e}^n -  \eta \nabla_{\vartheta_{t,e}^n} f_n(\vartheta_{t,e}^n,\mathbf{X}_{t,e}^{n})$;
            \ENDFOR    
    	\ENDFOR
        \STATE Aggregate selected local models to obtain: 
        \STATE ${\vartheta}_{t} = \frac{1}{\mathcal{N}}\sum_{n=1}^{\mathcal{N}}\vartheta_{t}^n$;
    \ENDFOR
    \STATE Set trained global model $\vartheta = \vartheta_{T_1+T_2}$;

\end{algorithmic}
\end{algorithm}

Motivated by the above findings, we propose a novel and fine-grained framework named \Name which is illustrated in Figure~\ref{fig:pipeline}, where the detailed workflow is shown in Algorithm~\ref{alg:1}. Our method \Name has the following innovations.

First, considering that different samples on client have a different degree of drift to the local updates, we evaluate and sort the loss set $\mathcal{S}_{n}$ of all the samples on $n$th client by global model from the server as follows:

\begin{equation}\label{eq:loss}
\mathcal{S}_{n}=\left\{\left(\boldsymbol{x}_{i}^{n},S_{k}^{n}\right)\right\}_{i=1}^{N_{n}}, \quad \text{with } S_{i}^{n} < S_{j}^{n} \quad \text{if } i < j
\end{equation}
where $\mathcal{S}_{n}$ is the evalution loss set of $n$-th client and $S_{k}^{n}$ denotes the bottom $k$th loss in all sample from $n$th client. Second, in order to divide all sample into biased samples and unbiased sample from $n$th client, we need to set a classification
sample $S_{\text{mid}}^{n}$  which when the loss value $S_{i}^{n} \leq S_{\text{mid}}^{n}$, $x_{k}^{n}$ sample corresponding to the loss value $S_{i}^{n}$ is unbiased sample while the rest is biased samples. However, as mentioned above, different rounds, different clients, and different samples will lead to different thresholds of classification points. Thus, we introduce the uncertainty to adaptively set the classification point for all clients on each round:

\begin{equation}\label{eq:uncertainty}
{\alpha}({{x}_{i}^{n}};\vartheta)) = 1 -(\textbf{max}({p}({{x}_{i}^{n}};\vartheta)) - \textbf{min}({p}({{x}_{i}^{n}};\vartheta)))
\end{equation}

where ${\alpha}({{x}_{i}^{n}};\vartheta))$ denotes the uncertainty of the model $\vartheta$ for the sample ${{x}_{i}^{n}}$ on $n$th client.

We plot the relationship between loss and uncertainty in the model's assessment of the sample as Figure \ref{fig:uncertainty}. 
With the increase of sample loss value, the model's perception of the sample is essentially divided into three stages from figure. Specifically, the model's judgment of sample is certain and accurate, the model's judgment of sample is uncertain and inaccurate, and the model's judgment of sample is certain but inaccurate. Based on the above properties, we can adaptively divide the sample into biased and unbiased sample by the highest uncertainty. Therefore, we take the sample with the highest uncertainty as the classification point at the beginning of client model training:

\begin{equation}\label{eq:classification}
{\alpha}(S_{\text{mid}}^{n})=\textbf{max}({\alpha}({{x}_{i}^{n}};\vartheta)))\end{equation}

Then, we propose progressively adding biased samples along during local training process in a epoch-by-epoch manner as follows:
\begin{equation}\label{eq:progressive}
\mathbf{X}_{t,e}^{n}=\mathbf{X}_{no}^{n}+\alpha \mathbf{X}_{bias}^{n}, \quad \alpha=\frac{1-\cos \left(\frac{e}{e_{\text {total }}} \pi\right)}{2}
\end{equation} where $\mathbf{X}_{t,e}^{n}$ denotes the ${n}$th client training sample on client epoch $\text {e}$ and global communication round $\text {t}$, $\mathbf{X}_{no}$ is the unbiased sample on ${n}$th client, $\mathbf{X}_{bias}$ is biased samples on ${n}$th client, $E_{\text {e }}$ is current client training epoch and $E_{\text {total }}$ is total epoch in the client training. Note that these samples are in the global round communication on the ${n}$th client.

One detail is that we do not adopt the scheme of linearly increasing biased samples epoch by epoch, such as $\mathbf{X}_{t,e}^{n}=\mathbf{X}_{no}^{n}+\frac{e}{e_{\text {total }}} \mathbf{X}_{bias}^{n}$. The specific comparison results are in section \ref{sec:discussion}. Concretely, we can discover that the samples near the classification point are relatively dense, while the samples at the edge of loss are relatively scattered, and the span of loss is large from figure \ref{fig:uncertaintysample}. This may result in the linear increment method not fitting the sample curve well. More discussion is demonstrated in section \ref{sec:discussion}.

In addition to the above, we also need an extra first stage as the warmup. This is because the model's initial judgment of the point of uncertainty fluctuates greatly. Figure \ref{fig:fluctuate} has shown the point. The intuition behind this is that the initial knowledge of the model is not enough, and it cannot make a relatively stable judgment on the client sample. Thus in order to ensure the stability of the initial training stage, warmup is used as the cold start stage of the model, which is to enable the model to acquire more diversified knowledge at the initial stage. Note that Figure~\ref{fig:pipeline} only describes the second stage.

In summary, our algorithm is divided into two stages: the first stage is Diversified knowledge acquisition stage, at which the client normally selects all the samples. And the second stage is Progressive knowledge learning stage. At this time, the algorithm first divides adaptive samples according to our above description, and then progressively adds unbiased samples during local client training.

\section{Experiments}\label{exp}

\subsection{Experiment Setting} \label{experiment_settings}

\textbf{Datasets and models.} We evaluate the performance of the proposed \Name over two models and four mainstream datasets. In Non-IID setting with label distribution skew, we consider Fashion-MNIST~\citep{DBLP:journals/corr/abs-1708-07747}, CIFAR-10~\citep{krizhevsky2009learning} and CIFAR-100~\citep{krizhevsky2009learning}, which contains 10, 10, 100 classes respectively. For CIFAR-10 and CIFAR-100 datasets, we use ResNet-50~\citep{he_deep_2016} as the backbone to train and test the performance while for Fashion-MNIST we use a simple CNN instead. The simple CNN has two 5x5 convolution layers (the first with 32 channels, the second with 64, each followed with 3x3 max pooling), a fully connected layer with 512 units and ReLU activation. 
In Non-IID setting with feature distribution skew, we utilize DomainNet dataset.

\textbf{Data Partition.} Follow the setting ~\citep{mcmahan14advances}, we adopt two classic data partitioning strategies, namely label distribution skew and feature distribution skew.
\begin{itemize}
\item \textbf{Label Distribution Skew:} To evaluate the performance of our work in a heterogeneous scenario with label distribution skew, we specify two Non-IID data partition methods called Shards~\citep{DBLP:conf/aistats/McMahanMRHA17} and Dirichlet~\citep{lin2020ensemble}. In the Shards setting, the sorted samples are shuffled into $ N*S $ shards, and assigned to $N$ clients randomly. Each client owns an equal number of pieces. In the second setting, data distribution over clients satisfies the Dirichlet distribution by using $\alpha$ to characterize the degree of heterogeneity. We set $\alpha$ of Dirichlet: $\{0.1, 0.3, 0.5\}$ and shards for each client: $\{2, 4, 8\}$.
\item \textbf{Feature Distribution Skew:}  In this setting, clients share the same label space while different feature distribution, which has been extensively studied in previous works~\citep{li2021fedbn,peng2019moment,yang2023fedfed}. We conduct the classification task on natural images sourced from DomainNet~\citep{peng2019moment}, which consists of diverse distributions of images from six distinct data sources. we utilize all six distinct data domains. We selected ten categories—airplane, clock, axe, ball, bicycle, bird, strawberry, flower, pizza, and bracelet—for the classification task. Each client operates exclusively with the data from one domain, and in this setting, all six clients participated in the aggregation of the model.
\end{itemize}

\textbf{Baselines.} Beside of FedAvg, we also compare against various types of Non-IID federated learning approaches with the proposed method in our experiments. The first main type includes typical non-aggregation methods that calibrate local update, including Scaffold~\citep{pmlr-v119-karimireddy20a}, FedProx~\citep{DBLP:conf/mlsys/LiSZSTS20}, FedExP~\citep{DBLP:conf/iclr/Jhunjhunwala0J23},FedLC~\citep{pmlr-v162-zhang22p} and FedRS~\citep{10.1145/3447548.3467254}. Besides, another representative strategy is to select aggregation on the sever, such as InCo~\citep{chan2024internal} and FedCDA~\citep{wang2024fedcda}.

\textbf{Implementation Details.} We implement the whole experiment in a simulation environment based on RTX 3090 GPUs. We use 100 clients in total and randomly choose 10\% each round for local training. We set the local epoch to 10, batch size to 64, and learning rate to $1e-3$. We employ SGD optimizer with momentum of $1e-4$
and weight decay of $1e-5$ for all methods and datasets. At the same time, we set the number of global communication rounds to $200$. For evaluation, we compute the average accuracy and standard deviation over the final 10 rounds of each run. For our method, we set the number of warmup rounds to $50$.

\subsection{Result with Label Distribution Skew}

\begin{table*}[ht]
\centering
\resizebox{\linewidth}{!}{
\begin{tabular}{l| c c c| c c c| c c c}
\toprule[1pt] 
 Method  & \multicolumn{3}{c}{Fashion-MNIST(\%)} 
 & \multicolumn{3}{c}{CIFAR-10(\%)} 
 & \multicolumn{3}{c}{CIFAR-100(\%)}\\

\midrule
\emph{Shards} ($S$) 
    & 2 & 4 & 8
    & 2 & 4 & 8
    & 2 & 4 & 8\\
\midrule
FedAvg &62.68\scriptsize{±6.30} &
67.31\scriptsize{±4.10} &
73.64\scriptsize{±2.16} &
23.56\scriptsize{±3.54} & 32.17\scriptsize{±2.52} & \textbf{36.18}\scriptsize{±2.25} & 2.99\scriptsize{±0.26} & 4.95\scriptsize{±0.37} & 7.95\scriptsize{±0.76} \\
FedProx &63.42\scriptsize{±5.23} &
68.71\scriptsize{±3.55} &
72.58\scriptsize{±3.54} &
20.65\scriptsize{±1.84} & 27.18\scriptsize{±3.13} & 31.89\scriptsize{±1.17} & 2.34\scriptsize{±0.21} & 4.47\scriptsize{±0.48} & 6.30\scriptsize{±0.62} \\
FedExP &54.84\scriptsize{±4.80} &
65.58\scriptsize{±3.54} &
70.19\scriptsize{±4.22} &
21.74\scriptsize{±2.74} & 29.90\scriptsize{±1.83} & 35.86\scriptsize{±2.39} & 2.71\scriptsize{±0.21} & 5.08\scriptsize{±0.68} & 8.40\scriptsize{±0.42} \\
FedLC &64.67\scriptsize{±0.25} &
74.20\scriptsize{±0.39} &
76.64\scriptsize{±0.44} &
21.28\scriptsize{±0.23} & 25.19\scriptsize{±0.35} & 32.07\scriptsize{±0.27} & 4.33\scriptsize{±0.09} & 7.06\scriptsize{±0.11} & 9.18\scriptsize{±0.12} \\
FedRS &62.17\scriptsize{±0.25} &
66.23\scriptsize{±0.69} &
68.64\scriptsize{±0.27} &
19.28\scriptsize{±0.13} & 23.47\scriptsize{±0.17} & 28.66\scriptsize{±0.32} & 4.01\scriptsize{±0.07} & 7.32\scriptsize{±0.10} & 8.80\scriptsize{±0.08} \\
Scaffold &65.72\scriptsize{±1.11} &
75.16\scriptsize{±1.33} &
\textbf{82.51}\scriptsize{±0.99} &
23.37\scriptsize{±0.35} & 27.08\scriptsize{±5.67} & 31.52\scriptsize{±2.67} & 5.01\scriptsize{±0.24} & 6.20\scriptsize{±0.12} & 8.18\scriptsize{±0.32} \\
FedCDA &66.47\scriptsize{±0.12} &
74.81\scriptsize{±0.17} &
80.54\scriptsize{±0.09} &
22.33\scriptsize{±0.12} & 31.38\scriptsize{±0.21} & 36.04\scriptsize{±2.34} & 5.52\scriptsize{±0.13} & 5.06\scriptsize{±0.32} & 7.51\scriptsize{±0.57} \\
InCo &65.66\scriptsize{±4.85} &
73.32\scriptsize{±2.09} &
75.68\scriptsize{±1.53} &
23.21\scriptsize{±0.54} & 29.94\scriptsize{±0.70} & 31.12\scriptsize{±0.65} & 6.02\scriptsize{±0.07} & 9.11\scriptsize{±0.24} & 10.11\scriptsize{±0.15} \\
Ours &\textbf{67.10}\scriptsize{±0.07} &
\textbf{76.01}\scriptsize{±0.33} &
79.31\scriptsize{±0.87} &
\textbf{25.97}\scriptsize{±0.33} & \textbf{32.81}\scriptsize{±0.28} & 34.04\scriptsize{±0.23} & \textbf{6.21}\scriptsize{±0.13} & \textbf{10.45}\scriptsize{±0.26} & \textbf{10.96}\scriptsize{±0.31} \\

\midrule
\emph{Dirichlet} ($\alpha$) & 0.1 & 0.3 & 0.5 & 0.1 & 0.3 & 0.5 & 0.1 & 0.3 & 0.5\\
\midrule
FedAvg &69.48\scriptsize{±2.59} &
74.50\scriptsize{±2.07} &
76.41\scriptsize{±0.94} &
27.72\scriptsize{±1.73} & 35.07\scriptsize{±1.80} & 37.88\scriptsize{±1.58} & 11.79\scriptsize{±0.34} & 14.00\scriptsize{±0.35} & 14.50\scriptsize{±0.18}\\
FedProx &61.39\scriptsize{±6.17} &
70.65\scriptsize{±3.38} &
73.87\scriptsize{±2.37} &
23.36\scriptsize{±2.57} & 26.86\scriptsize{±2.92} & 30.16\scriptsize{±2.47} & 7.99\scriptsize{±0.55} & 10.70\scriptsize{±0.33} & 11.03\scriptsize{±0.18}\\
FedExP &68.08\scriptsize{±4.56} &
74.29\scriptsize{±2.11} &
77.14\scriptsize{±1.17} &
27.78\scriptsize{±3.44} & 35.00\scriptsize{±1.93} & 39.06\scriptsize{±1.24} & 11.61\scriptsize{±0.52} & 13.53\scriptsize{±0.29} & 14.54\scriptsize{±0.25}\\
FedLC &67.11\scriptsize{±0.31} &
69.27\scriptsize{±0.16} &
74.19\scriptsize{±1.09} &
27.99\scriptsize{±0.28} & 32.42\scriptsize{±1.25} & 40.76\scriptsize{±1.02} & 11.88\scriptsize{±0.26} & 13.00\scriptsize{±0.12} & 14.17\scriptsize{±0.58} \\
FedRS &62.67\scriptsize{±0.25} &
64.20\scriptsize{±0.09} &
68.64\scriptsize{±0.86} &
26.34\scriptsize{±0.23} & 30.47\scriptsize{±0.17} & 38.07\scriptsize{±0.67} & 14.02\scriptsize{±0.12} & 15.06\scriptsize{±0.11} & 16.77\scriptsize{±0.37} \\
Scaffold &73.43\scriptsize{±1.76} &
74.95\scriptsize{±0.76} &
76.96\scriptsize{±0.69} &
14.54\scriptsize{±2.22} & 18.38\scriptsize{±1.43} & 19.26\scriptsize{±2.63} & 12.96\scriptsize{±0.18} & 13.20\scriptsize{±0.26} & 14.20\scriptsize{±0.40} \\
FedCDA &70.66\scriptsize{±4.85} &
71.11\scriptsize{±1.34} &
73.68\scriptsize{±1.24} &
29.21\scriptsize{±0.15} & 38.94\scriptsize{±0.34} & 39.68\scriptsize{±0.04} & 12.34\scriptsize{±0.42} & 14.81\scriptsize{±0.14} & 15.44\scriptsize{±0.16} \\
InCo &69.66\scriptsize{±1.13} &
73.32\scriptsize{±2.09} &
75.68\scriptsize{±1.53} &
28.21\scriptsize{±4.55} & 36.94\scriptsize{±2.71} & 38.69\scriptsize{±1.65} & 11.52\scriptsize{±0.67} & 13.81\scriptsize{±0.24} & 14.22\scriptsize{±0.15} \\
Ours &\textbf{76.24}\scriptsize{±0.16} &
\textbf{80.34}\scriptsize{±0.15} &
\textbf{82.11}\scriptsize{±0.17} &
\textbf{38.49}\scriptsize{±0.22} & \textbf{43.11}\scriptsize{±0.25} & \textbf{45.27}\scriptsize{±0.19} & \textbf{18.37}\scriptsize{±0.27} & \textbf{19.11}\scriptsize{±0.12} & \textbf{19.08}\scriptsize{±0.11}\\

\bottomrule[1pt]
\end{tabular}
}
\caption{The comparison of test accuracy of different methods. The best results are \textbf{bolded}.}
\label{tab:heterogeneity}
\end{table*}
\par

We report the comparison results with other baselines in Table \ref{tab:heterogeneity}. In order to demonstrate the generalization of our method, we compare them on two different Non-IID settings, Shards and Dirichlet distribution. We apply different data distributions on different datasets. We can see that our proposed \Name achieves the best performance on almost all settings. 
It demonstrates the effectiveness and benefit of \Name. Specifically, on relatively larger datasets such as CIFAR-10 Dirichlet 0.5, \Name with ResNet-50 achieves 45.27\% accuracy whereas the best baseline method FedLC achieves 40.76\% accuracy. In addition, \Name with simple CNN also makes improvements on relatively smaller datasets, and the improvement is not less than in large models. At the same time, we can also see that the results of our method on relatively small datasets and simple CNN are not the best, which may be because the client biased samples of different rounds is less drift to the global model on small datasets and simple models, and can not provide better performance of the global model by progressive learning. In conclusion, we can notice our \Name makes more improvements on the large model, complex datasets than small model, simple datasets. 

\begin{table*}

\centering
\resizebox{\linewidth}{!}{
\begin{tabular}{l| c c c| c c c| c c c}
\toprule[1pt] 
 Method  & \multicolumn{3}{c}{Fashion-MNIST(\%)} 
 & \multicolumn{3}{c}{CIFAR-10(\%)} 
 & \multicolumn{3}{c}{CIFAR-100(\%)}\\

\midrule
Noisy Label Radio& 0.1 & 0.3 & 0.5 & 0.1 & 0.3 & 0.5 & 0.1 & 0.3 & 0.5\\
\midrule
FedAvg &80.19\scriptsize{±0.13} &
78.44\scriptsize{±0.11} &
75.28\scriptsize{±0.31} &
42.15\scriptsize{±0.27} &
36.78\scriptsize{±0.42} & 28.23\scriptsize{±0.35} & 13.70\scriptsize{±0.16} & 11.13\scriptsize{±0.23} & 8.08\scriptsize{±0.16}\\
FedProx &79.94\scriptsize{±0.10} &
78.17\scriptsize{±0.15} &
75.60\scriptsize{±0.27} &
37.96\scriptsize{±0.24} & 32.67\scriptsize{±0.20} & 26.30\scriptsize{±0.53} & 10.76\scriptsize{±0.27} & 8.75\scriptsize{±0.15} & 6.39\scriptsize{±0.14}\\
FedExP &80.27\scriptsize{±0.11} &
78.28\scriptsize{±0.13} &
75.74\scriptsize{±0.25} &
41.82\scriptsize{±0.18} & 36.60\scriptsize{±0.24} & 29.32\scriptsize{±0.50} & 13.44\scriptsize{±0.23} & 11.11\scriptsize{±0.18} & 7.51\scriptsize{±0.24}\\
FedLC &80.11\scriptsize{±0.11} &
78.18\scriptsize{±0.09} &
76.98\scriptsize{±0.12} &
39.23\scriptsize{±0.26} & 35.14\scriptsize{±0.30} & 30.41\scriptsize{±0.18} & 15.08\scriptsize{±0.39} & 11.34\scriptsize{±0.21} & 10.39\scriptsize{±0.17} \\
FedRS &79.86\scriptsize{±0.22} &
77.59\scriptsize{±0.11} &
74.88\scriptsize{±0.13} &
37.88\scriptsize{±0.24} & 34.28\scriptsize{±0.25} & 27.68\scriptsize{±0.38} & 12.09\scriptsize{±0.18} & 10.09\scriptsize{±0.17} & 9.21\scriptsize{±0.28} \\
Scaffold &78.65\scriptsize{±0.09} &
78.88\scriptsize{±0.09} &
76.40\scriptsize{±0.17} &
40.72\scriptsize{±0.36} & 37.14\scriptsize{±0.19} & 29.36\scriptsize{±0.41} & 12.37\scriptsize{±0.24} & 9.94\scriptsize{±0.15} & 7.39\scriptsize{±0.06} \\
FedCDA &79.63\scriptsize{±0.15} &
78.20\scriptsize{±0.12} &
75.74\scriptsize{±0.30} &
39.83\scriptsize{±0.15} & 35.27\scriptsize{±0.21} & 30.49\scriptsize{±0.28} & 14.18\scriptsize{±0.22} & 10.33\scriptsize{±0.21} & 9.10\scriptsize{±0.21} \\
InCo &79.89\scriptsize{±0.56} &
77.27\scriptsize{±0.60} &
75.16\scriptsize{±0.46} &
38.12\scriptsize{±1.13} & 34.55\scriptsize{±0.99} & 24.78\scriptsize{±1.75} & 11.02\scriptsize{±0.34} & 9.54\scriptsize{±0.29} & 8.28\scriptsize{±0.54} \\
Ours &\textbf{80.63}\scriptsize{±0.14} &
\textbf{79.91}\scriptsize{±0.13} &
\textbf{78.88}\scriptsize{±0.08} &
\textbf{44.46}\scriptsize{±0.22} & \textbf{40.27}\scriptsize{±0.29} & \textbf{38.96}\scriptsize{±0.19} & \textbf{16.01}\scriptsize{±0.17} & \textbf{15.47}\scriptsize{±0.28} & \textbf{12.13}\scriptsize{±0.34}\\

\bottomrule[1pt]
\end{tabular}}
\caption{The comparison of test accuracy on noisy label datasets. The best results are \textbf{bolded}.}
\label{tab:noise}
\end{table*}
\begin{figure}
    \centering
    \includegraphics[width=0.8 \linewidth]{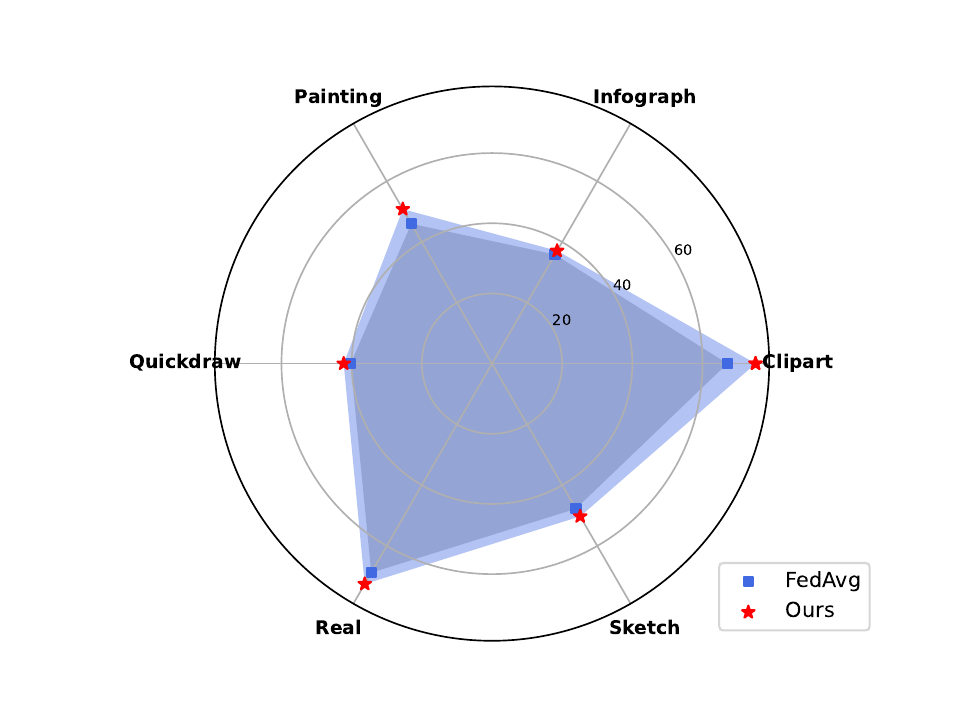}
    \caption{Result on DomainNet dataset}
\label{fig:domainnet}
\vspace{-10pt}
\end{figure}
Note that the baseline approaches we compare can actually be divided into two categories. One is calibrating local update, and the other is setting aggregating strategy. The performance of InCo in the table is not good enough. The main reason is that its perception of the client is relatively limited, so the improvement is small. Furtherly, we can see that scaffold outperforms our method \Name on some datasets, while it has not the good performance on some other datasets. This indicates that Scaffold is not stable and always sensitive to datasets and hyperparameters. The performance of our method \Name is usually more prominent when there is high heterogeneity setting. This may be because the biased samples drifts the model to a large extent when there is high heterogeneity. Therefore, our method can more effectively improve its performance in such scenarios. Besides, our experimental results show that the server aggregating strategy-based methods also improve less on small models and simple datasets. We consider that the main reason is that small models have little parameter, which the effect of directly improving the model parameters is not obvious in this scenario.

\subsection{Result with Feature Distribution Skew} In the above experiments, we define data heterogeneity with label distribution skew. In this section, we test our approach on feature Non-IID setting with feature distribution skew. We utilize domainNet dataset which contains six domains: clipart, infograph, painting, quickdraw, real, and sketch. We selected ten categories for the classification task and ResNet-18 as backbone. Each client operates exclusively with the data from one domain, and in this setting, all six clients participated in the aggregation of the model. Figure~\ref{fig:domainnet} is the result. We can find that our approach \Name outperforms FedAvg's performance not only on six very different domains, but also on the final global model.

\subsection{Result on Noise Label Datasets}
In real scenarios, data labels are often noisy or absent. In this section, we employ \Name on noise label dataset to test the robustness of our method. Table \ref{tab:noise} 
shows results of our method \Name. We set different noise label radio which is 0.1, 0.3 and 0.5 on three datasets. We can find that our method achieve better performance compared with other Non-IID methods. Specifically, our approach offers a significant improvement over other approaches for both relatively smaller dataset Fashion-MNIST and relatively larger datasets CIFAR-100. With the increase of noise label ratio, our method improves more and more significantly, while \Name can still maintain good performance even under the most severe noise label ratio condition which is 0.5. It proves that our approach is robust and scalable.

\subsection{More discussion and experiments}\label{sec:discussion}
\begin{figure}
    \centering
    \subfloat[Warmup Stage]{
    \label{fig:warmuprounds}
    \includegraphics[width=0.48\linewidth]{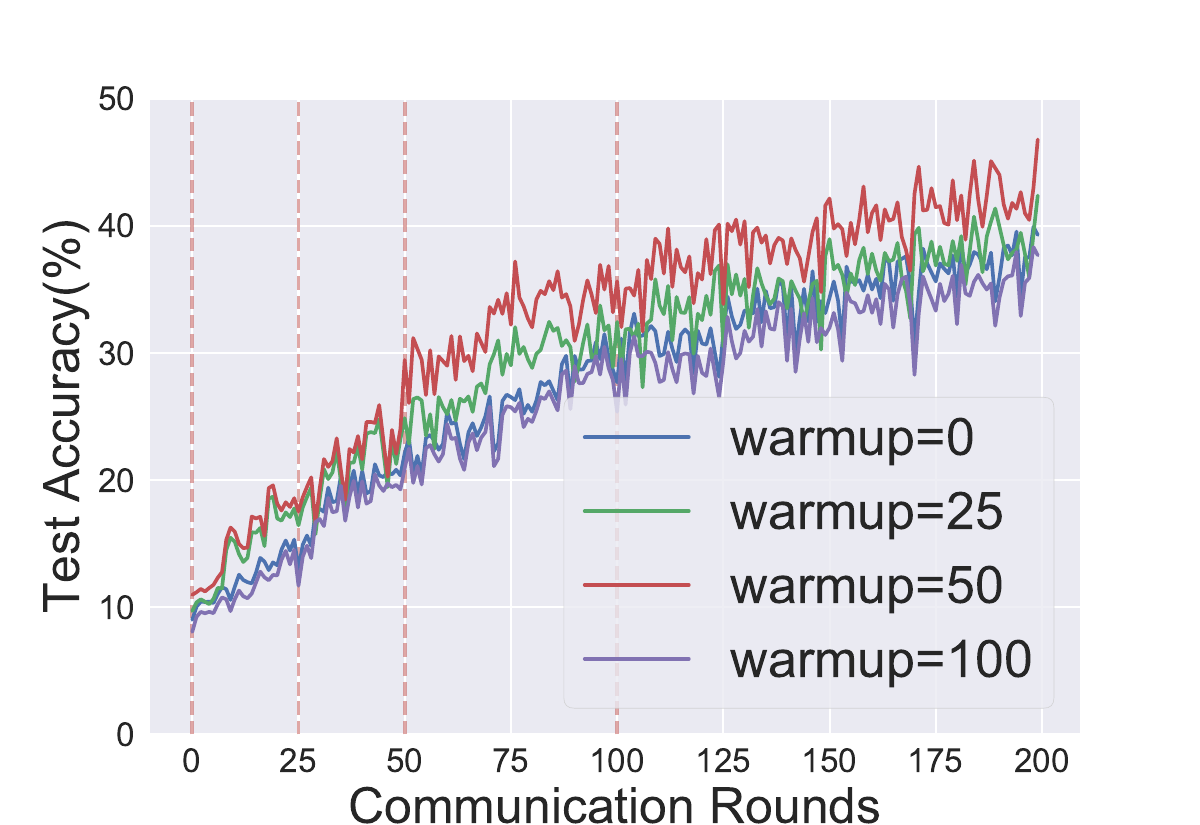}
    }
    \subfloat[Sample Selection Strategy]{
    \label{fig:agg}
    \includegraphics[width=0.48\linewidth]
    {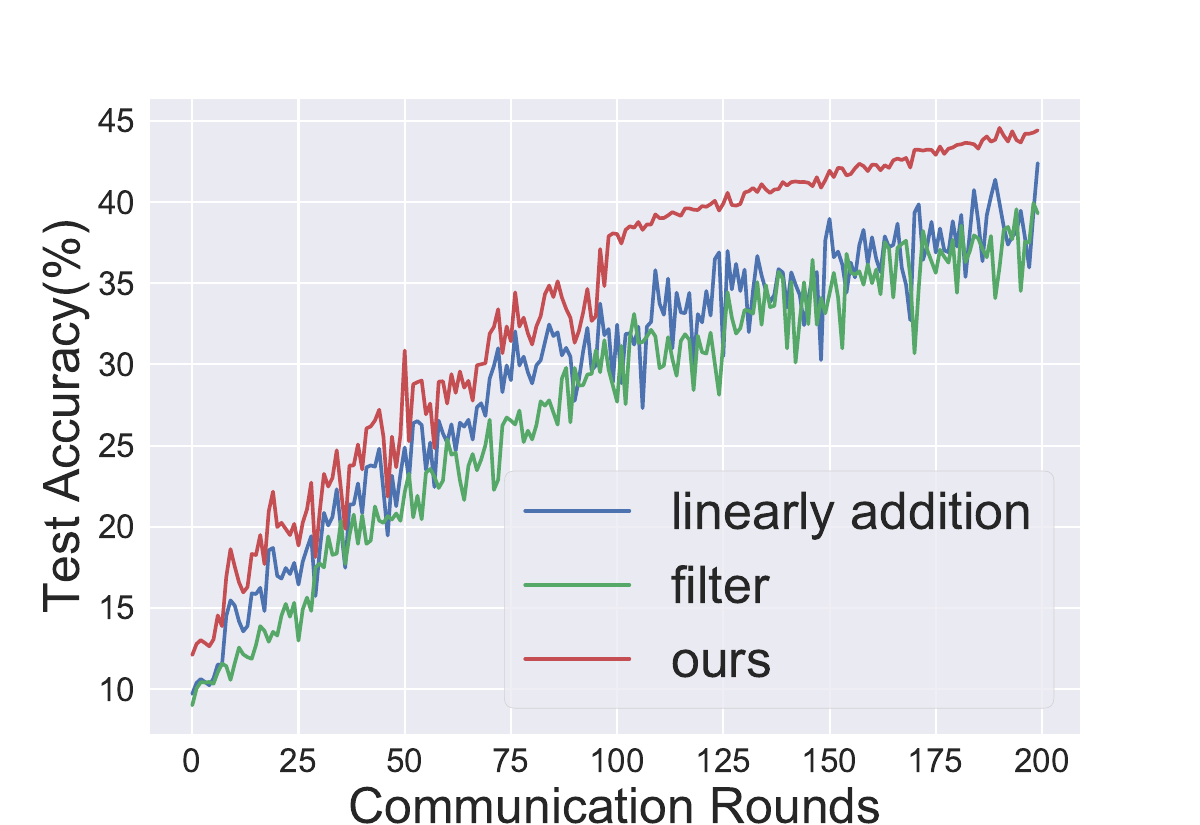}
    }
    \caption{Ablation Study}
\label{hyperparam}
\vspace{-10pt}
\end{figure}
In this section, we discuss two parts which are mentioned on section \ref{sec:motivation} in our method \Name. The first one is about first warmup stage. The second one is about sample selection strategy.

\textbf{Warmup Stage}
For the first one, we compare with three baselines. Our method sets the warmup rounds as $50$. We compare with the baseline which varies the warmup rounds as $\{0, 25, 100\}$. In fact, we show the result as Figure \ref{fig:warmuprounds}. From this, we can see that our method which sets the warmup rounds as 50. Note that this does not mean that 50 rounds is always optimal. However, we can discover that when the warmup round is zero which means the first warmup stage doesn't exist, the performance of the model is always worse. This may be due to the lack of diversity knowledge in the first stage. Certainly, when warmup rounds are too high, the performance is worse too. This is because the first stage is too long, resulting in too many global model drifts, which makes the progressive learning effect of the second stage not good. Therefore, the warmup rounds are a relatively sensitive parameter. We think the first warmup phase is a good trade-off between minimizing client drift and minimizing knowledge sacrification. Specifically, the first phase helps the model acquire diversified knowledge, while the second phase mitigated the phenomenon of client draft.

\textbf{Sample Selection Strategy} We have discussed the biased sample selection strategy in section \ref{sec:motivation}. To summarize, there are three strategies: filter, linearly addition and our method. We compared these three strategies on Cifar10 dataset and use ResNet-18 as the backbone while other parameters are the same as above. Figure \ref{fig:agg} shows the result. Our method proves the superiority and it outperforms the other two strategies. It proves that filtering biased samples may cause the model to lose part of its knowledge and fall into local optimal solutions or slow convergence. On the other hand, the effect of linear inclusion of biased sample is not good, which may be due to the sample distribution, which is more concentrated in the middle part and dispersed in the edge part. This may result in a linear function that does not fit the sample distribution well.

\section{Conclusion}\label{sec:conclusion}
This paper focuses on Non-IID setting in federated learning. We have observed that there exists substantial bias in each sample and the client drift can be viewed as a cumulative manifestation of the biases of all samples. Besides, the bias between samples is different from each other. Motivated by this finding, we propose a novel two-stage effective method for FL called \Name that exploits the drift properties of different samples. Our method is to adopt a bias-aware sample selection scheme that dynamically selects the samples from small biases to large epoch by epoch to train progressively the local model in each round. We analyze and demonstrate its effectiveness and robustness through a lot of experiments.

\section{Acknowledgments}
This work was supported by the National Natural Science Foundation of China (NSFC Grant No. 62402331), the Fundamental Research Funds for the Central Universities (Grant No. YJ202429), the Fundamental Research Funds for the Central Universities ( No. SCU2024D012), and the Science and Engineering Connotation Development Project of Sichuan University (No. 2020SCUNG129).


\bibliography{aaai25}

\end{document}